\documentclass[twoside,11pt]{article}
\usepackage{blindtext}

\usepackage{jmlr2e}
\usepackage{algorithm}
\usepackage{algpseudocode}
\usepackage{booktabs}
\usepackage{amsmath}

\usepackage{lastpage}
\jmlrheading{X}{2025}{1-\pageref{LastPage}}{X/25; Revised X/X}{X/X}{21-0000}{Carlos Velez Garcia, Miguel Cazorla and Jorge Pomares}

\ShortHeadings{Planning as Descent}{Velez-Garcia et al.}
\firstpageno{1}

\begin{document}

\title{Planning as Descent: Goal-Conditioned Latent Trajectory Synthesis in Learned Energy Landscapes}

\author{\name Carlos Vélez García \email cvelez@inescop.es \\
       \addr Robotics \& Automation \\
       INESCOP\\
       Elda, Alicante, Spain
       \AND
       \name Miguel Cazorla \email miguel.cazorla@ua.es \\
       \addr University Institute for Computing Research\\
       University of Alicante\\
       Alicante, E03690, Spain
       \AND
       \name Jorge Pomares \email jpomares@ua.es \\
       \addr University Institute for Computing Research\\
       University of Alicante\\
       Alicante, E03690, Spain}

\editor{}

\maketitle

\begin{abstract}%   <- trailing '%' for backward compatibility of .sty file

We present Planning as Descent (PaD), a framework for offline goal-conditioned reinforcement learning that grounds trajectory synthesis in verification. Instead of learning a policy or explicit planner, PaD learns a goal-conditioned energy function over entire latent trajectories, assigning low energy to feasible, goal-consistent futures. Planning is realized as gradient-based refinement in this energy landscape, using identical computation during training and inference to reduce train-test mismatch common in decoupled modeling pipelines.

PaD is trained via self-supervised hindsight goal relabeling, shaping the energy landscape around the planning dynamics. At inference, multiple trajectory candidates are refined under different temporal hypotheses, and low-energy plans balancing feasibility and efficiency are selected.

We evaluate PaD on OGBench cube manipulation tasks. When trained on narrow expert demonstrations, PaD achieves state-of-the-art 95\% success, strongly outperforming prior methods that peak at 68\%. Remarkably, training on noisy, suboptimal data further improves success and plan efficiency, highlighting the benefits of verification-driven planning. Our results suggest learning to evaluate and refine trajectories provides a robust alternative to direct policy learning for offline, reward-free planning.

\end{abstract}


\section{Introduction}

Learning to act primarily from observation alone remains a central challenge in modern artificial intelligence \citep{lecun2022path}. This challenge is particularly acute in real-world domains such as robotics, where interaction is expensive, unsafe, or impractical, and where available data mostly consist of offline, reward-free trajectories collected under unknown and potentially suboptimal policies. In such settings, agents must infer how to achieve user-specified goals purely from heterogeneous demonstrations, without access to online exploration or reward signals.

We study this problem in the setting of offline goal-conditioned reinforcement learning (GCRL), where the objective is to reach arbitrary target states using only a static dataset of reward-free trajectories. Offline GCRL poses several fundamental difficulties: (i) extracting meaningful structure from unstructured and suboptimal data; (ii) composing disjoint behavioral fragments that may not co-occur within a single trajectory; (iii) propagating sparse goal information over long horizons; and (iv) reasoning about multi-modal futures under stochastic dynamics. Recent benchmarks such as OGBench \citep{park2024ogbench} highlight the difficulty of these challenges and show that many existing methods struggle to generalize robustly to \emph{unseen} goals.

A common strategy for addressing offline decision making is to separate modeling and planning. Model-based methods learn forward dynamics and then perform trajectory optimization or model predictive control (MPC) at inference time \citep{zhou2024dino, hansen2023td, sobal2025learning}. While conceptually appealing, this separation often leads to train--test mismatches: powerful optimizers can exploit small inaccuracies in learned dynamics models, producing adversarial or physically implausible trajectories that fail at deployment time \citep{henaff2019model}. 

An alternative line of work reframes control as trajectory generation, using sequence models such as Decision Transformers \citep{chen2021decision}, masked trajectory models \citep{wu2023masked, janner2021offline, carroll2022uni}, or diffusion-based policies \citep{chi2023diffusion, janner2022planning}. These models directly model the distribution of trajectories and can synthesize diverse, multimodal behaviors from offline datasets. However, their sampling-based nature often leads to reproducing undesirable behaviors when trained on noisy or suboptimal data, and they lack explicit mechanisms for enforcing long-horizon dynamical feasibility or goal satisfaction. More broadly, these approaches learn \emph{how to generate trajectories}, but do not explicitly learn \emph{how to evaluate or verify them}\citep{west2023generative}.

In this work, we propose Planning as Descent (PaD), a framework that rethinks offline goal-conditioned control through the lens of generation by verification. Rather than learning a policy, generator, or explicit planner, PaD learns a goal-conditioned energy landscape over entire future trajectories. This energy assigns low values to trajectories that are dynamically plausible and consistent with a desired goal, and high values to incompatible ones. Planning then arises implicitly as gradient descent in this learned energy landscape, iteratively refining candidate trajectories to minimize their energy.

Crucially, PaD explicitly enforces alignment between training and inference by using the same gradient-based refinement procedure in both phases. The energy landscape is shaped during training around the exact descent dynamics used at test time, ensuring that inference corresponds to optimization behavior the model has been trained to support. The forward pass of the model computes trajectory energies (verification), while the backward pass provides structured descent directions that refine trajectories toward feasible, goal-consistent futures (synthesis). This \emph{training-as-inference} alignment helps mitigate the train--test discrepancies that arise in decoupled modeling and planning pipelines, and contrasts sharply with diffusion and masked trajectory models, which rely on stochastic sampling or reconstruction objectives. In PaD, planning is goal-directed and realized entirely through energy minimization, without autoregressive rollouts, learned noise schedules or value backups.

Moreover, our method is motivated by the most fundamental design bias in deep learning: \emph{depth enables composition}. Just as deep networks compose pixels into objects and words into sentences, PaD allows primitive transitions to compose into subgoals and plans within the depth of a single-learned energy-based model (EBM). Hierarchical planning structure occurs implicitly through gradient-based refinement in the representation space, potentially eliminating the need for explicitly engineered handcrafted abstractions. 

We demonstrate that PaD performs robustly across qualitatively different data regimes, including narrow expert demonstrations and broad, highly suboptimal datasets. On challenging tasks from the OGBench single-cube manipulation suite, PaD achieves state-of-the-art performance, exhibits strong robustness to distribution shift, and--perhaps counterintui\-tively--produces more efficient plans when trained on diverse but highly suboptimal data. These results suggest that learning to verify trajectories, rather than directly generating them, provides a powerful foundation for offline goal-conditioned planning. An implementation of the proposed method will be made publicly available at: \url{https://github.com/inescopresearch/pad}

\paragraph{Contributions.}
Our main contributions are:
\begin{itemize}
\item We introduce an energy-based formulation of offline GCRL that unifies trajectory evaluation and synthesis, casting planning as energy minimization in latent trajectory space.
\item We propose a self-supervised training scheme based on hindsight goal relabeling that aligns training-time verification with inference-time planning.
\item We present a gradient-based planning procedure that iteratively refines noisy latent trajectories into coherent, goal-directed plans.
\item We demonstrate state-of-the-art performance on OGBench single-cube tasks and provide an analysis showing that the learned energy landscape supports effective planning and generalization to unseen goals.
\end{itemize}

\section{Related Work}

We next review relevant previous work in model-based control, sequence modeling, and energy-based approaches, and position our contribution in this landscape.

\subsection{Offline Goal-Conditioned Reinforcement Learning}
Offline Goal-Conditioned Reinforcement Learning (GCRL) studies goal-reaching from reward-free offline datasets, typically using hindsight relabeling \citep{andrychowicz2017hindsight} to enable unsupervised training. Early goal-conditioned behavioral cloning methods (GCBC) \citep{lynch2020learning, ghosh2019learning} perform well when state coverage is sufficient but degrade sharply under distribution shift or narrow expert demonstrations. Subsequent approaches such as GCIVL, GCIQL, QRL, CRL, and HIQL incorporate value learning, contrastive learning, or explicit hierarchical structure to better propagate sparse goal information and improve compositionality \citep{kostrikov2021offline, park2023hiql, eysenbach2022contrastive, wang2023optimal}. However, these methods remain sensitive to dataset quality and often perform well in either narrow or broad noisy regimes, but not both \citep{park2024ogbench}.

A unifying characteristic of these methods is that they learn an explicit policy or value function, with planning either implicit in policy execution or mediated through value-based rollouts. This coupling exposes them to well-known offline RL failure modes, such as extrapolation and compounding action errors at inference time \citep{levine2020offline}.

Planning as Descent (PaD) departs from this paradigm by avoiding policy and value learning altogether; instead, it learns a goal-conditioned energy function over entire future trajectories. Planning arises through gradient-based refinement in this energy landscape, providing a verification-driven alternative that enables robust goal-reaching across diverse offline data regimes.

\subsection{Model-Based Planning and Trajectory Optimization.}
Model-based approaches address offline control by learning a forward dynamics model and performing planning through trajectory optimization or model predictive control (MPC) \citep{rawlings2020model, luo2024survey}. Recent methods \citep{zhou2024dino, hansen2023td, sobal2025learning} leverage latent dynamics models and planning in representation space to improve scalability and robustness. However, decoupling model learning from planning often leads to a train-test mismatch, as powerful optimizers can exploit small modeling errors and produce physically implausible or brittle plans at inference time \citep{talvitie2017self}.

Several works attempt to mitigate this issue through uncertainty regularization \citep{henaff2019model}, short-horizon planning, or frequent replanning, but still rely on explicit forward rollouts through learned dynamics. As a result, planning quality remains tightly coupled to model accuracy, particularly in long-horizon and distribution-shifted settings.

Planning as Descent (PaD) avoids explicit dynamics modeling and forward simulation altogether. Instead, PaD learns a goal-conditioned energy function over entire future trajectories and performs planning directly via gradient-based refinement in trajectory space. By unifying trajectory evaluation and synthesis within a single learned energy landscape, PaD sidesteps model exploitation and reduces train-test mismatch inherent to model-based planning pipelines.

\subsection{Trajectory Generation via Sequence Modeling}
An alternative to model-based planning reframes control as a trajectory generation problem, using sequence models trained on offline data. Decision Transformers \citep{chen2021decision} condition autoregressive models on goals or returns to synthesize action sequences \citep{schmidhuber2019reinforcement}, while masked trajectory models reconstruct missing segments of trajectories using bidirectional context \citep{wu2023masked, janner2021offline, carroll2022uni}. More recently, diffusion-based policies and planners \citep{janner2022planning, chi2023diffusion} model the distribution of expert trajectories and have achieved state-of-the-art results through imitation learning on large-scale foundational models \citep{black2024pi_0, liu2024rdt, bjorck2025gr00t}. 

Despite their success, these approaches fundamentally learn how to generate trajectories that resemble the training distribution \citep{west2023generative}. As a result, they are prone to reproducing suboptimal or undesirable behaviors present in the data and often rely on sampling procedures, noise schedules, or autoregressive rollouts that introduce variance and compounding errors over long horizons. Moreover, these methods lack an explicit mechanism for evaluating or verifying whether a synthesized trajectory is dynamically feasible or goal-consistent \citep{balim2025model}.

Planning as Descent (PaD) departs from generative trajectory modeling by learning a goal-conditioned energy function over entire future trajectories. Rather than sampling trajectories, PaD refines latent candidate future plans via gradient descent in the learned energy landscape, explicitly optimizing for feasibility and goal satisfaction through verification rather than likelihood or reconstruction. 

\subsection{Energy-Based Models for Control and Planning}
Energy-Based Models (EBMs) learn an explicit energy function whose low-energy configurations correspond to compatible system states or actions \citep{lecun2006tutorial, du2019implicit}. In control settings, EBMs have been used to model conditional action distributions, most notably in Implicit Behavior Cloning (IBC), where actions are obtained at inference time by minimizing a learned energy landscape \citep{liu2020energy, florence2022implicit, davies2025ebt}. However, most EBM-based control methods rely on contrastive training objectives that require large numbers of negative samples, often resulting in sharp or poorly conditioned energy landscapes that make inference-time optimization difficult and unstable. This limitation has historically hindered the scalability and practical deployment of EBMs in long-horizon control and planning tasks.

Recent work has revisited EBM training from an optimization-based perspective, introducing objectives that implicitly regularize the energy landscape and avoid explicit negative sampling \citep{wang2023energy, gladstone2025energy}. While these advances significantly improve training stability and scalability, their potential for addressing core challenges world modeling, control and planning--such as long-horizon reasoning and goal-conditioned trajectory synthesis--remains largely unexplored.

In contrast, PaD uses a single goal-conditioned energy function directly as a planner, where gradient-based minimization over entire latent trajectories constitutes the planning procedure itself. Action decoding is decoupled from trajectory planning and can be learned with substantially fewer labeled samples, as it only needs to model local state-to-action mappings rather than long-horizon planning behavior.

\section{Background}

\subsection{Offline Goal-Conditioned RL}

We consider a reward-free Markov decision process (MDP) $ \mathcal{M} = ( \mathcal{S}, \mathcal{A}, \mu, p)$, where $\mathcal{S}$ is the state space, $\mathcal{A}$ is the action space, $\mu \in \mathcal{P}(\mathcal{S})$ is the initial-state distribution, and $ p(s' \mid s,a)$ denotes the (possibly stochastic) transition dynamics kernel specifying the probability of transitioning to state $s' \in S$ after taking action $a \in A$ from state $s \in S$. 

We study this problem in the \emph{offline} setting, where the agent has access only to a fixed, reward-free dataset $\mathcal{D} = \{\tau^{(n)}\}_{n=1}^{N} $ of state-action trajectories $\tau^{(n)} =(s^{(n)}_{0}, a^{(n)}_{0}, s^{(n)}_{1}, \ldots, s^{(n)}_{T_{n}})$ collected by an unknown behavior policy. Importantly, these trajectories are \emph{not necessarily optimal}, reflecting the heterogeneous, partially exploratory, and potentially multi-modal nature of real-world data. 

The objective of offline goal-conditioned RL is to learn a \emph{goal-conditioned policy} $\pi:\mathcal{S}\times{S} \rightarrow \mathcal{A}$ that enables the agent to reach any target state $ s_g \in \mathcal{S} $ from any initial state $ s_0 \in \mathcal{S} $ in as few steps as possible. We define the sparse goal-reaching reward: $ r_g(s) = \mathbb{I}[s = s_g] $ 
and seek a policy that maximizes the expected discounted return
\[
J(\pi) = \mathbb{E}_{\pi, p} \Bigg[ \sum_{t=0}^{T} \gamma^{t} r_g(s_t) \,\Bigg|\, s_0 \sim \mu, s_g \sim \mathcal{S} \Bigg],
\]
where $\gamma \in (0,1]$ is the discount factor. 

A key property of this formulation is that the \emph{entire state space} serves as the goal space. A goal is specified by a full state, not by a subset of features such as object positions. This choice yields a fully unsupervised objective, enabling domain-agnostic training from unlabeled, reward-free data \citep{park2024ogbench}. 

\subsection{Energy-Based Models}

Energy-Based Models (EBMs) provide a flexible framework for modeling dependencies between variables by associating a scalar \emph{energy}—a measure of compatibility—to each configuration of variables \citep{lecun2006tutorial}. 
Let $x \in \mathcal{X}$ denote an observed variable and $y \in \mathcal{Y}$ denote a target variable. 
An EBM defines an \emph{energy function}
\[
E_{\theta}: \mathcal{X} \times \mathcal{Y} \rightarrow \mathbb{R},
\]
parameterized by $\theta$, assigning a scalar energy $E_{\theta}(x,y)$ to every pair $(x,y)$.

\paragraph{Inference.} 
Given $x$, inference consists of finding the most compatible configuration of $y$ by minimizing the energy:
\[
y^{*} = \arg\min_{y \in \mathcal{Y}} E_{\theta}(x, y).
\]

\paragraph{Learning.} 
Learning consists of shaping the energy landscape so that \emph{compatible} pairs $(x,y)$ lie in low-energy regions and \textit{incompatible} pairs lie in high-energy regions. 
In practice, this is done by minimizing the energy of positive (observed) pairs while maximizing the energy of negative (corrupted) pairs, typically through contrastive or regularized training procedures that avoid computing the generally intractable partition function (i.e., the normalization constant $Z_\theta(x) = \int_y{\exp({-E_\theta(x, y')})dy'}$ which is prohibitive to compute in high-dimensional spaces). Following \citep{wang2023energy}, we adopt an optimization-based perspective on EBM training that implicitly regularizes the energy landscape and enables scalable learning. 

\section{Method}
\label{sec:method}
We introduce \emph{Planning as Descent} (PaD), a goal-conditioned latent-space planning framework in which future trajectories are synthesized by descending a learned conditional energy landscape. The core idea is to learn an energy function over latent trajectories that simultaneously serves as a \emph{verifier} of dynamical plausibility and goal satisfaction, and as a \emph{planner} whose gradient field specifies how candidate trajectories should be refined. Given past observations and a desired goal specification, the model assigns low energy to latent futures that are feasible and goal-consistent, while the gradient of this energy provides the descent direction required to synthesize such trajectories.

\begin{figure}[!htb]
    \centering
    \includegraphics[width=\textwidth]{"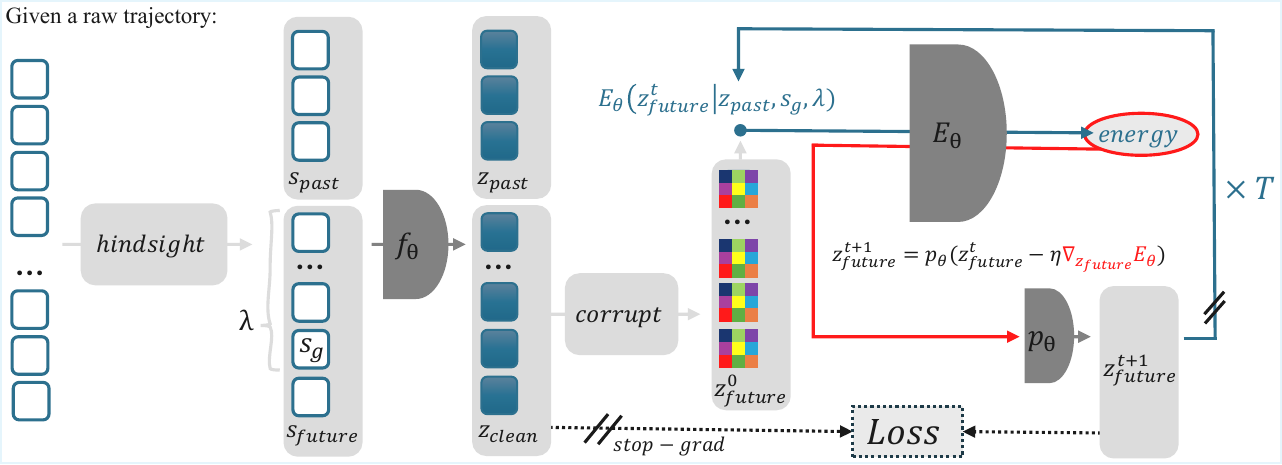"}
    \caption{\textbf{Overview of the Planning as Descent (PaD) learning framework.} 
Given trajectory states and a hindsight-relabeled goal $(s_g, \lambda)$, states are independently encoded into latent representations using $f_\theta$. 
Future latents are corrupted to form an initial trajectory $z_{\mathrm{future}}^{0}$, which is iteratively refined by descending the conditional energy $E_\theta$ and projecting updates back onto the encoder-induced manifold through $p_\theta$. 
At each refinement step, a denoising loss compares the intermediate trajectory to the clean future latents while stop-gradient operations prevent (i) mode collapse and (ii) backpropagation through the refinement dynamics.}
    \label{fig:method}
\end{figure}

PaD consists of three main components trained jointly end-to-end: (i) a state encoder $f_\theta$ that maps individual observations into a latent space without modeling temporal structure; (ii) a conditional energy function $E_\theta$ that jointly evaluates and refines entire latent trajectories; and (iii) a projector network $p_\theta$ that ensures refinement remains confined to the encoder-induced manifold. A defining property of PaD is that the same refinement mechanism is used during both training (Algorithm~\ref{alg:training}) and inference (Algorithm~\ref{alg:inference}), allowing the energy landscape to be shaped around the planning dynamics.

\subsection{Latent State Representation}

Given an observation sequence $(s_0,\ldots,s_T)$, each state $s_t$ is encoded independently as a latent vector $z_t = f_\theta(s_t)$, where the encoder $f_\theta : \mathcal{S} \rightarrow \mathbb{R}^d$ captures only state-wise information and does not impose temporal dependencies. Given past latents $z_{\mathrm{past}} = (z_0, \ldots, z_k)$ and a goal specification $(s_g,\lambda)$, where $\lambda \in [0,1]$ is a normalized time-to-reach variable specifying the desired relative point within the planning horizon at which the goal should be reached, PaD introduces the future latent sequence
\[
    z_{\mathrm{future}} = (z_{k+1},\ldots,z_{k+H}),
\]
which serves as a free optimization variable during planning. Importantly, $\lambda$ does not correspond to a physical time or fixed number of environment steps, but instead provides a relative temporal conditioning signal that allows the planner to reason jointly about trajectory feasibility and goal-reaching speed.

\subsection{Conditional Energy Model and Gradient-Based Planning}

We define a conditional energy
\[
    E_\theta(z_{\mathrm{future}} \mid z_{\mathrm{past}}, s_g, \lambda),
\]
which assigns a scalar value to a candidate future trajectory $z_{\mathrm{future}}$, conditioned on the latent past $z_{\mathrm{past}}$, the goal state $s_g$, and the continuous time-to-reach parameter $\lambda \in [0,1]$. Low energy indicates that the trajectory is dynamically plausible and consistent with reaching the goal within the temporal budget encoded by~$\lambda$.

PaD performs planning by iteratively refining the latent trajectory through gradient descent on this energy landscape. Given a current trajectory estimate $z_{\mathrm{future}}^{(t)}$, the raw refinement step is
\[
    z_{\mathrm{future-raw}}^{(t)}
    =
    z_{\mathrm{future}}^{(t)}
    -
    \eta \,\nabla_{z_{\mathrm{future}}}
    E_\theta\!\left(
        z_{\mathrm{future}}^{(t)} \,\middle|\, z_{\mathrm{past}}, s_g, \lambda
    \right),
\]
where $\eta$ denotes the refinement step size.

To ensure that refinement remains on or near the encoder-induced manifold, the updated trajectory is passed through a shallow learnable projector $p_\theta$, producing the final update rule:
\begin{equation}
\label{eq:update_rule}
    z_{\mathrm{future}}^{(t+1)}
    =
    p_\theta\!\left(
        z_{\mathrm{future}}^{(t)}
        -
        \eta \,\nabla_{z_{\mathrm{future}}}
        E_\theta\!\left(
            z_{\mathrm{future}}^{(t)} \,\middle|\, z_{\mathrm{past}}, s_g, \lambda
        \right)
    \right).
\end{equation}

\begin{figure}[!htb]
    \centering
\includegraphics[width=0.6\textwidth]{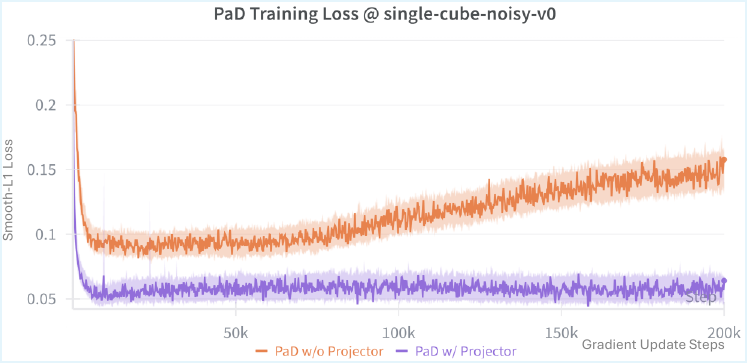}
\caption{Training loss on \texttt{single-cube-noisy-v0} when ablating the
manifold projector. The lightweight 130K-parameter projector substantially
stabilizes training and improves convergence with negligible computational
overhead.}
\label{fig:projector_ablation}
\end{figure}

\paragraph{The projector matters.}
We empirically find that the projector $p_\theta$ plays a critical role
in stabilizing refinement. Figure~\ref{fig:projector_ablation}
shows that, in the absence of the projector, training becomes unstable,
as energy gradients alone may push latent states toward off-manifold
regions that do not correspond to valid encoded observations, leading
to degenerate latents and degraded planning performance.

Conversely, the projector cannot produce meaningful refinements in the
absence of the structured descent directions supplied by the energy model.
Effective refinement therefore arises from the interaction between the
two components: the energy function provides informed, goal-conditioned
descent directions, while the projector enforces representational validity
by mapping refined trajectories back onto the encoder-induced manifold.
We further ablate the projector component in Section~\ref{sec:projector_ablation}.

\subsection{Training as Inference}

A defining property of PaD is that the refinement procedure used for planning is identical during training and inference.  
This “training-as-inference’’ principle regularizes the energy landscape such that its gradient flow implements the desired planning behavior.

\paragraph{Hindsight goal relabeling with temporal targets.}
\label{par:hindsight}
Given a trajectory of length $L$, we first sample a scalar $r \in [0,1]$ from the truncated arccos distribution $p(r)=\frac{2}{\pi}(1-r^{2})^{-1/2}$ and map it linearly to a past-window length $P_{\mathrm{past}} \in [1,P_{\max}]$, where $P_{\max}$ is the maximum allowed past context.  
This biases training toward larger past windows while remaining aligned with inference-time conditions.

To specify the temporal target, we draw $\lambda \sim \mathcal{U}(0,1)$ and map it linearly to a future index $G \in [P_{\max}, H]$, where $H$ is the planning horizon.  
The state at index $G$ becomes the goal $s_g$.  
Conditioning on $\lambda$ is essential: without an explicit temporal target, the model would treat all trajectories that eventually reach the goal as equivalent, making denoising from heavily corrupted latents considerably more difficult.

\paragraph{Latent trajectory corruption.}
Let $z_{\mathrm{clean}}$ denote the clean future latent trajectory $f_\theta(s_{\mathrm{future}})$.  
To mimic uncertainty over future predictions, we take a corruption scheme inspired by diffusion models  
\[
    z_{\mathrm{future}}^{(0)}
    =
    \sqrt{\beta}\, z_{\mathrm{clean}}
    +
    \sqrt{1 - \beta}\,\epsilon,
    \qquad
    \epsilon \sim \mathcal{N}(0,I)
\]
with the corruption level $\beta$ uniformly sampled as $\beta \sim \mathcal{U}(0,1)$.
\paragraph{Denoising-based training objective.}
Starting from this noisy initialization, the model performs $T$ refinement steps using Equation~\ref{eq:update_rule}.  
At each refinement step, the intermediate trajectory $z_{\mathrm{future}}^{(t)}$ is compared to the clean target $z_{\mathrm{clean}}$ using a smooth-$L_1$ distance.  
The target is treated as a constant, and stop-gradient is applied to $z_{\mathrm{future}}^{(t)}$ between refinement steps to prevent backpropagation through the refinement dynamics themselves, which empirically accelerates training and improves stability.

The overall training loss is therefore
\[
    \mathcal{L}
    =
    \sum_{t=1}^{T}
    \ell\!\left( z_{\mathrm{future}}^{(t)},\, z_{\mathrm{clean}} \right).
\]

Importantly, this loss is backpropagated through the entire optimization process, which requires second-order derivatives--specifically, gradients of gradients with respect to model parameters arising from the refinement steps. These second-order terms are computed efficiently as Hessian-vector products, which increases training cost about $1.66\times$ compared to standard first-order backpropagation in a feed-forward model, assuming a single refinement step and all other factors held constant \citep{gladstone2025energy}.

We summarize the training procedure of the proposed framework in Algorithm~\ref{alg:training}.

\begin{algorithm}[H]
\caption{\textbf{Training}: Latent Trajectory Denoising with Hindsight Goal Relabeling.}
\label{alg:training}
\textbf{Inputs:} clean trajectory $s_{0:L}$, encoder $f_\theta$, planner $E_\theta$, projector $p_\theta$, refinement steps $T$
\begin{algorithmic}[1]

\State $(s_{\mathrm{past}},\, s_{\mathrm{future}}, s_g, \lambda)
        \gets$ \textsc{Hindsight}$(s_{0:L})$ 
        \Comment{Hindsight goal relabeling (See \ref{par:hindsight})}

\State $z_{\mathrm{past}} \gets f_\theta(s_{\mathrm{past}})$ 
        \Comment{Encode past states}

\State $z_{\mathrm{clean}} \gets \textsc{StopGradient}               (f_\theta(s_{\mathrm{future}}))$ \Comment{Prevent representation collapse}

\State $\beta \sim \mathcal{U}(0,1)$ \Comment{Sample corruption level}
\State $\epsilon \sim \mathcal{N}(0,I)$ \Comment{Sample Gaussian noise}

\State 
$z_{\mathrm{future}} \gets 
\sqrt{\beta}\, z_{\mathrm{clean}} 
+ \sqrt{1-\beta}\, \epsilon$ 
\Comment{Corrupt future latents}

\State $\mathcal{L} \gets 0$

\For{$t = 1$ to $T$}

    \State $E \gets E_\theta(z_{\mathrm{future}}, z_{\mathrm{past}}, s_g, \lambda)$ 
    \Comment{Compute energy}

    \State $z_{\mathrm{future}} \gets z_{\mathrm{future}} - \eta \nabla_{z_{\mathrm{future}}} E$ 
    \Comment{Gradient refinement}

    \State $z_{\mathrm{future}} \gets p_\theta(z_{\mathrm{future}})$ 
    \Comment{Projection}

    \State $\mathcal{L} \gets \mathcal{L} + \text{smooth-$L_1$}(z_{\mathrm{future}}, z_{\mathrm{clean}})$
    \Comment{Accumulate denoising loss}

    \State $z_{\mathrm{future}} \gets \textsc{StopGradient}(z_{\mathrm{future}})$ \Comment{Prevent backprop-through-time}

\EndFor

\State Update $\theta$ using $\mathcal{L}$

\end{algorithmic}
\end{algorithm}

\subsection{Inference with Multi-Hypothesis Temporal Targets}
\label{subsec:inference}

\begin{figure}[!htb]
    \centering
    \includegraphics[width=\textwidth]{"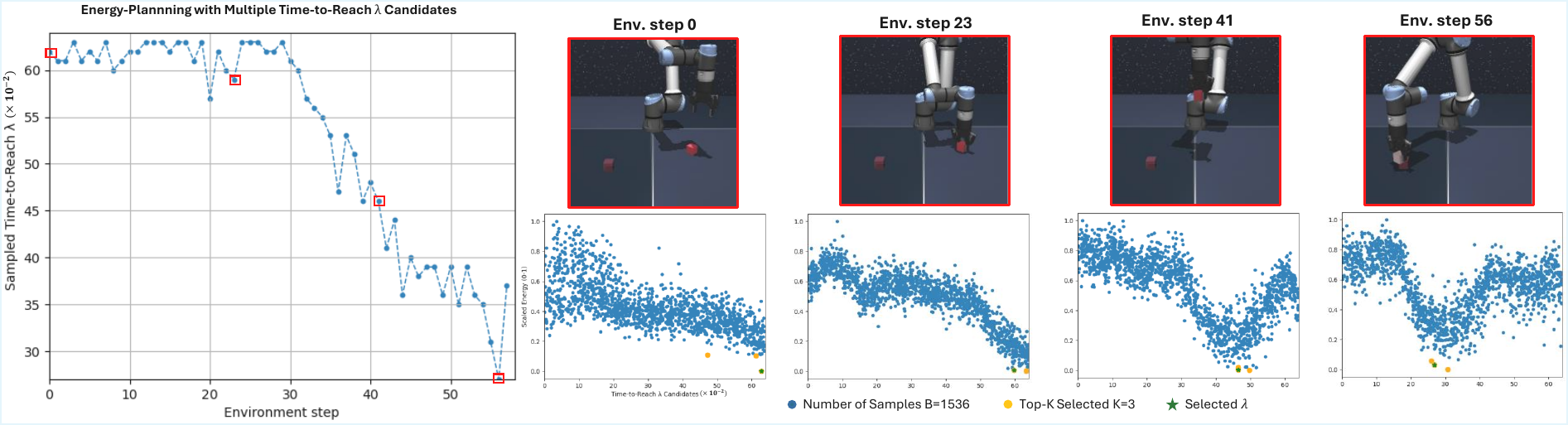"}
    \caption{\textbf{Multi-hypothesis planning and temporal target selection in PaD.} \textbf{Left:} As task execution progresses, PaD naturally selects time-to-reach hypotheses ($\lambda$) that correspond to decreasing distances to the goal, reflecting adaptive planning as the agent approaches completion. \textbf{Right:} Example timesteps during the rollout. The top row displays the corresponding environment states at each step. The bottom row shows the distribution of sampled temporal targets and their associated energies, with lower energies indicating more plausible plans for reaching the goal.}
    \label{fig:PaD_planning}
\end{figure}

At test time, PaD only has access to the sequence of past observations $s_{\mathrm{past}} = (s_0,\ldots,s_L)$ and a desired goal state $s_g$.  
Inference proceeds by sampling and refining multiple candidate future trajectories under different temporal hypotheses as shown in Figure \ref{fig:PaD_planning}.  

At test time, PaD is provided with a sequence of past observations $s_{\mathrm{past}} = (s_0, \ldots, s_k)$ and a desired goal state $s_g$. Inference proceeds by synthesizing and refining multiple candidate future trajectories under different normalized time-to-reach hypotheses $\lambda$, allowing the model to jointly reason about \emph{how} to reach the goal and \emph{how quickly} it should be reached.

Given the encoded past $z_{\mathrm{past}} = f_\theta(s_\mathrm{past})$, we initialize a batch of $B$ candidate future trajectories  
\[
    z_{\mathrm{future}}^{(0)} \sim \mathcal{N}(0,I)^{B},
\]
each paired with an independently sampled temporal hypothesis $\lambda_b \sim \mathcal{U}(0,1)$. As in training, $\lambda$ serves as a relative temporal conditioning signal rather than an absolute number of environment steps, and conditions the planner on the intended fraction of the planning horizon at which the goal should be achieved.
Every candidate is refined for $T$ steps using the same update rule as in training (Eq.~\ref{eq:update_rule}).

After refinement, each trajectory is scored by its final energy
\[
    E_b = E_\theta\!\left(
        z_{\mathrm{future}}^{(T,b)}
        \,\middle|\,
        z_{\mathrm{past}}, s_g, \lambda_b
    \right),
\qquad b=1,\ldots,B,
\]
where superscript $(T,b)$ denotes the $b$-th candidate after $T$ refinement steps. Lower energies correspond to trajectories that are both dynamically plausible and consistent with reaching the goal at the rate specified by $\lambda_b$.

PaD selects a final plan in two stages. First, it identifies the $K$ candidates with the lowest energies. Second, it samples one trajectory from this top-$K$ set using a categorical distribution with logits proportional to $-\lambda$, introducing a mild bias toward plans that reach the goal sooner whenever doing so remains energetically feasible. This selection mechanism--summarized in Algorithm~\ref{alg:inference}--balances feasibility and efficiency within the learned energy landscape and mirrors the temporal conditioning used during training.

\begin{algorithm}[H]
\caption{\textbf{Inference}: Energy Planning with Multiple Time-to-Reach $\lambda$ Candidates.}
\label{alg:inference}

\textbf{Inputs:} past sequence $s_{\mathrm{past}}$, goal state $s_g$, encoder $f_\theta$, planner $E_\theta$, 
projector $p_\theta$, refinement steps $T$, 
number of samples $B$, top-$k$ set size $K$

\begin{algorithmic}[1]

\State $z_{\mathrm{past}} \gets \mathrm{tile}\left(f_\theta \left( s_{\mathrm{past}}\right), B\right)$ \Comment{Encode past states}
\State $z_{\mathrm{future}} \sim \mathcal{N}(0,I)^B$ \Comment{Sample $B$ future latents}
\State $\lambda \sim \mathcal{U}(0,1)^B$ \Comment{Sample $B$ Steps-to-reach candidates}

\For{$t = 1$ to $T$}
    \State $E \gets E_\theta(z_{\mathrm{future}}, z_{\mathrm{past}}, s_g, \lambda)$ 
    \Comment{Compute energies}

    \State $z_{\mathrm{future}} \gets z_{\mathrm{future}} - \eta \nabla_{z_{\mathrm{future}}} E$ 
    \Comment{Gradient refinement}

    \State $z_{\mathrm{future}} \gets p_\theta(z_{\mathrm{future}})$ 
    \Comment{Projection}
\EndFor

\State $E \gets E_\theta(z_{\mathrm{future}}, z_{\mathrm{past}}, s_g, \lambda)$ 
\Comment{Compute final energies}

\State $(E_{\text{top}},\, \mathcal{I}_K) \gets \textsc{TopK}(-E, K)$ 
\Comment{Select $K$ lowest-energy plans}

\State $i_{\mathrm{final}} \sim \mathrm{Categorical}(\mathrm{logits} = -\lambda_{\mathcal{I}_K})$ 
\Comment{Sample index biased by lower Steps-to-reach}

\State $z_{\mathrm{plan}} \gets z_{\mathrm{future}}[i_{\mathrm{final}}]$

\State \Return $z_{\mathrm{plan}}$

\end{algorithmic}
\end{algorithm}

More advanced inference schemes (e.g., proposal distributions conditioned on previous refinements) are possible but intentionally omitted to highlight the intrinsic capability of the learned energy landscape. We leave such extensions to future work.

\paragraph{Online Replanning.}
At inference, PaD is deployed within an iterative replanning loop in which planning and execution alternate. After synthesizing a latent future trajectory using the refinement procedure described above, only the first $N$ predicted transitions are decoded into actions via the inverse dynamics model $g_\psi$ and executed open-loop in the environment. The agent then receives new observations, updates the past window, resamples temporal hypotheses, and invokes the same refinement mechanism to obtain a fresh plan. This plan–execute–replan structure enables PaD to continually correct plan inaccuracies and stochastic transitions while remaining computationally lightweight, as all refinement steps are fully parallelized across hypotheses. In Section~\ref{subsec:ablation-N}, we ablate the choice of the replanning interval $N$ and analyze how it affects task performance, stability, and computational cost.

\subsection{Action Decoding via Inverse Dynamics}

Once a latent future trajectory has been synthesized by PaD, it must be translated into executable control inputs. To this end, we employ a separate inverse dynamics model
\[
    a_t = g_\psi(z_t, z_{t+1}),
\]
which maps consecutive latent states to the action responsible for the corresponding transition.

The inverse dynamics model $g_\psi$ is trained independently from the planner using supervised learning on action-labeled transitions drawn from the offline dataset. Importantly, gradients from $g_\psi$ do \emph{not} propagate to the encoder or energy model; planning is therefore learned entirely from state-only trajectories, and action labels are not required for shaping the energy landscape or the refinement dynamics.

In our experiments, for simplicity, $g_\psi$ is trained using the full set of available action-labeled transitions. However, this choice is not fundamental to the method. Since inverse dynamics only needs to model local, single-step state transitions rather than long-horizon planning behavior, it can in principle be trained from a substantially smaller labeled subset without affecting the planner itself. We leave a systematic study of the trade-off between action-label availability and execution performance to future work.

Decoupling action decoding from latent planning provides two key advantages. First, it enables PaD to learn goal-conditioned planning entirely from action-free data, which is particularly relevant in settings where only state observations are available. Second, it prevents imperfections in the inverse dynamics model from distorting the learned planning representation, as planning quality is determined solely by the energy-based refinement in latent space.

\section{Experiments}
\label{sec:experiments}

We evaluate Planning as Descent (PaD) on two state-based manipulation tasks from the OGBench cube suite \citep{park2024ogbench}.  OGBench is a challenging multi-goal benchmark spanning diverse manipulation and locomotion tasks under both state and pixel observations. Compared to earlier offline suites such as D4RL \citep{fu2020d4rl}, which has become largely saturated as of 2025 \citep{park2025flow}, OGBench provides substantially more difficult evaluation scenarios and offers datasets with distinct state-coverage characteristics (e.g., narrow expert play vs.\ broad noisy demonstrations). These properties make it a suitable testbed for studying PaD under varying data distributions. 

\paragraph{Cube-single task details.} The \texttt{cube-single} task involves pick-and-place manipulation of cube blocks, where the goal is to control a robotic arm to move a cube into a target configuration. The robot is controlled at the end-effector level with a 5-dimensional action space, corresponding to displacements in the $x$, $y$, and $z$ positions, as well as gripper yaw, and gripper opening. Task success is determined solely by the final cube configuration; the arm pose itself is not considered when evaluating success. 

\begin{figure}[!htb]
    \centering
    \includegraphics[width=\textwidth]{"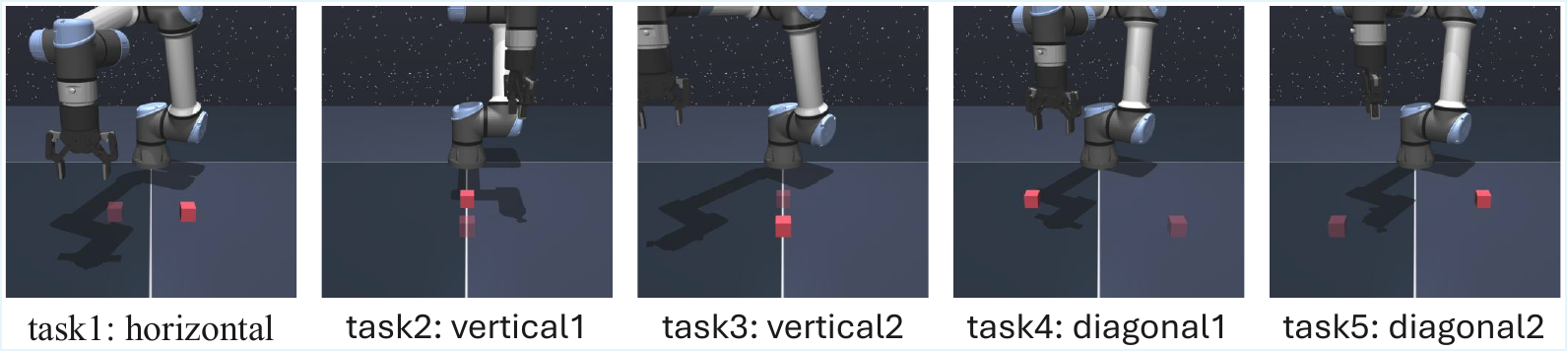"}
    \caption{\textbf{Cube-single tasks.}}
    \label{fig:tasks}
\end{figure}

At test time, each task corresponds to one of five movement types illustrated in Figure \ref{fig:tasks}. For each episode, both the initial and target cube poses are randomized within the constraints of the selected task movement type, ensuring diverse goal configurations during evaluation.

\paragraph{Evaluation protocol.}
We follow the official multi-goal evaluation structure of OGBench. Each policy is evaluated on the five test-time tasks, and for each task, we perform 20 randomized goal evaluation episodes, yielding 100 total rollouts per policy. Performance is measured using the binary success rate of \citet{park2024ogbench}. OGBench considers a score above 95 as indicative of successful goal achievement, providing a consistent threshold for evaluating agent performance across tasks. To account for training stochasticity, we independently train and evaluate eight policies with distinct random seeds, matching the multi-seed protocol used for state-based OGBench baselines. All results are reported as mean $\pm$ standard deviation across the eight runs. 

Unless otherwise stated, baseline results (GCBC, GCIVL, GCIQL, HIQL, CRL, QRL) are taken directly from the official OGBench benchmark \citep{park2024ogbench} and correspond to models trained and evaluated under the full benchmark protocol. In contrast, PaD is trained and evaluated under a reduced computational budget (200K training updates vs.\ 1M, a single evaluation checkpoint vs.\ three, and 20 episodes per goal vs.\ 50), making the comparison conservative with respect to PaD. Additional evaluation details are provided in Appendix~\ref{app:evaluation}. 

\paragraph{Implementation details.}
PaD is implemented using lightweight neural architectures. The state encoder $f_\theta$ is a two-layer MLP followed by LayerNorm, applied independently to each state without temporal mixing. The energy model $E_\theta$ consists of three stages: (i) a 2-layer 1D convolutional encoder that processes the concatenated past and future latent trajectories and reduces the temporal resolution by a factor of four; (ii) a decoder-only Transformer with three blocks, each configured with four attention heads and causal attention masks; and (iii) a final linear projection that maps the Transformer outputs to a scalar energy value. Goal conditioning is incorporated by appending two learned tokens—one corresponding to the goal state $s_g$ and one to the temporal target $\lambda$. The goal token is produced by a shallow three-layer MLP. For the temporal target, $\lambda$ is first encoded using a fixed sinusoidal embedding with logarithmically spaced frequencies, after which the resulting representation is passed through a three-layer MLP to produce the temporal token. The projector $p_\theta$ is implemented as a two-layer MLP. The inverse dynamics model $g_\psi$ is an MLP with two hidden layers and is trained separately using supervised learning on action-labeled transitions; importantly, gradients from $g_\psi$ do not propagate to the encoder, energy model, or projector, ensuring that planning is learned entirely independently of action decoding.

The full PaD model contains almost 6.5M trainable parameters, which is lightweight by contemporary standards. All PaD planning components (i.e., $f_\theta$, $E_\theta$, and $p_\theta$) are trained end-to-end for 200K gradient updates using the AdamW Optimizer with a batch size of 512 on a single NVIDIA RTX~3090 GPU, resulting in roughly 9 hours of wall-clock training time. The inverse dynamics model $g_\psi$ is trained separately under the same dataset but does not influence planner optimization. Unless otherwise specified, we adopt the following default hyperparameters throughout our main experiments: refinement steps $T = 2$, number of temporal hypotheses $B = 768$, top-$K$ candidate set size $K = 5$, and replanning interval $N = 1$.

\subsection{RQ1: Can PaD Learn from Narrow Expert Demonstrations?}
\label{sec:expert}
We evaluate PaD on the goal-conditioned manipulation task \texttt{cube-single-play-v0}.
The dataset contains 1M state transitions collected from open-loop scripted demonstrations with temporally correlated action noise. Although these demonstrations achieve consistently high success, they exhibit extremely narrow state coverage and provide no corrective behavior, resulting in a highly biased training distribution.

This experiment tests whether PaD can (i) extract meaningful planning structure from such narrow expert demonstrations and (ii) remain robust under inference-time distribution shift, where small deviations can push the planner outside the limited expert manifold.

\begin{table}[!htb]
\caption{\textbf{Success rates on \texttt{cube-single-play-v0}.} PaD substantially outperforms all baselines under narrow expert demonstrations by a large margin, achieving high success across tasks and an overall score at the OGBench 95-point success threshold.}
\label{tab:play}
\centering
\begin{tabular}{@{}llllllll@{}}
\toprule
Task              & GCBC      & GCIVL & GCIQL & QRL & CRL & HIQL & PaD \\ \midrule
task1 & 7$\pm$\small{3} & 57$\pm${6} & 71$\pm$\small{9} & 6$\pm$\small{2} &  20$\pm$\small{6} & 15$\pm$\small{5} & \textbf{94$\pm$\small{3}} \\
task2 & 5$\pm$\small{2} & 51$\pm${6} & 71$\pm$\small{6} & 5$\pm$\small{2} &  20$\pm$\small{4} & 16$\pm$\small{5} & \textbf{95$\pm$\small{2}} \\
task3 & 7$\pm$\small{3} & 55$\pm${6} & 70$\pm$\small{6} & 4$\pm$\small{1} &  21$\pm$\small{6} & 16$\pm$\small{3} & \textbf{98$\pm$\small{3}} \\
task4 & 4$\pm$\small{2} & 50$\pm${4} & 61$\pm$\small{8} & 4$\pm$\small{2} &  16$\pm$\small{3} & 14$\pm$\small{5} & \textbf{93$\pm$\small{4}} \\
task5 & 4$\pm$\small{2} & 52$\pm${6} & 67$\pm$\small{7} & 4$\pm$\small{3} &  15$\pm$\small{3} & 13$\pm$\small{4} & \textbf{95$\pm$\small{5}} \\ \midrule
\textbf{overall} & 6$\pm$\small{2} & 53$\pm$\small{4} & 68$\pm$\small{6} & 5$\pm$\small{1} & 19$\pm$\small{2} & 15$\pm$\small{3} & \textbf{95$\pm$\small{2}} \\ \bottomrule
\end{tabular}
\end{table}

\paragraph{Results} in Table \ref{tab:play} show that PaD achieves a remarkable overall success rate of 95$\pm$2, substantially outperforming all competing baselines. Across the five test tasks, PaD consistently achieves individual task success rates in the 93--98\% range, with the lowest being 93$\pm$4 and the highest 98$\pm$3. In comparison, the strongest baseline (GCIQL) achieves an overall success rate of 68$\pm$6, while GCIVL achieves 53$\pm$4, and other baselines like GCBC, QRL, CRL, and HIQL all remain below 20\%. The results highlight a consistently large gap between PaD and all other approaches, with standard deviations across seeds remaining low for PaD, indicating stable performance. Notably, these results are obtained despite PaD being trained for only 200K updates, compared to 1M updates used for the reported baselines.

\paragraph{Interpretation.} These results demonstrate that PaD can effectively leverage even narrowly distributed, expert-level demonstrations to achieve highly robust and generalizable planning performance. Despite being trained solely on demonstrations with limited state coverage and lacking corrective behaviors, PaD is able to synthesize coherent, goal-directed trajectories at test time and generalize to unseen goals. This suggests that the energy-based planning-by-descent framework is able to extract and recombine meaningful behavioral primitives from expert data, overcoming the limitations of direct imitation and static behavioral cloning. The strong and stable performance also indicates that PaD’s iterative refinement and planning mechanism confers robustness to distribution shift, allowing the model to recover from deviations and avoid cascading errors that typically hinder baselines under narrow data regimes.

\subsection{RQ2: Can PaD Learn from Highly Suboptimal Demonstrations?}
\label{sec:noisy}

We next evaluate PaD on the goal-conditioned manipulation task \texttt{cube-single-noisy-v0}. 
This dataset also contains 1M state transitions, collected using closed-loop Markovian policies perturbed by per-episode Gaussian action noise. 
In contrast to the expert dataset, these demonstrations provide broad state coverage but exhibit inconsistent and highly suboptimal behavior, resulting in a diverse yet noisy training distribution.

This setting examines whether PaD can (i) leverage wide but noisy coverage to improve generalization and (ii) maintain planning robustness in the presence of inconsistent demonstration behavior.

\begin{table}[!htb]
\caption{\textbf{Success rates on \texttt{cube-single-noisy-v0}.} PaD achieves consistently high performance across all tasks when trained on highly suboptimal, noisy demonstrations, outperforming most baselines by a large margin and matching the strongest competing method, with both surpassing the 95-point success threshold. Results are reported as mean $\pm$ standard deviation over eight seeds.}
\label{tab:noisy}
\centering
\begin{tabular}{@{}llllllll@{}}
\toprule
Task  & GCBC & GCIVL & GCIQL & QRL & CRL & HIQL & PaD \\ \midrule
task1 & 5$\pm$\small{3} & 71$\pm$\small{12} & \textbf{100$\pm$\small{1}} & 17$\pm$\small{12} & 39$\pm$\small{4} & 48$\pm$\small{6} &  \textbf{99$\pm$\small{2}}  \\
task2 & 7$\pm$\small{5} & 70$\pm$\small{11} & \textbf{100$\pm$\small{0}} & 23$\pm$\small{8} & 39$\pm$\small{7} & 39$\pm$\small{7} &  \textbf{99$\pm$\small{2}}   \\
task3 & 1$\pm$\small{1} & 67$\pm$\small{10} & \textbf{99$\pm$\small{1}} & 4$\pm$\small{3} & 36$\pm$\small{7} & 41$\pm$\small{7} &  \textbf{100$\pm$\small{0}}  \\
task4 & 16$\pm$\small{5} & 76$\pm$\small{10} & \textbf{98$\pm$\small{2}} & 47$\pm$\small{14} & 36$\pm$\small{4} & 36$\pm$\small{6} &  \textbf{94$\pm$\small{5}}  \\
task5 & 12$\pm$\small{5} & 70$\pm$\small{12} & \textbf{100$\pm$\small{0}} & 37$\pm$\small{9} & 42$\pm$\small{5} & 44$\pm$\small{9} &  \textbf{98$\pm$\small{4}}  \\ \midrule
\textbf{overall}  & 8$\pm$\small{3} & 71$\pm$\small{9} & \textbf{99$\pm$\small{1}} & 25$\pm$\small{6} & 38$\pm$\small{2} & 41$\pm$\small{6} &  \textbf{98$\pm$\small{2}} \\ \bottomrule
\end{tabular}
\end{table}

\paragraph{Results} in Table \ref{tab:noisy} show that PaD achieves an overall success rate of 98$\pm$2, reliably surpassing the OGBench success threshold. Across all individual tasks, PaD’s results range from 94$\pm$5 to 100$\pm$0, with most tasks meeting or exceeding the 95-point mark. In comparison, the best baseline (GCIQL) achieves 99$\pm$1 overall, while other baselines fall short of the threshold, reinforcing PaD's robust and consistent ability to achieve successful outcomes even when trained on highly suboptimal and inconsistent demonstration data. Notably, these results are obtained despite PaD being trained for only 200K updates, compared to 1M updates used for the reported baselines.

\paragraph{Interpretation.} These results indicate that PaD is capable of robustly learning from datasets characterized by high diversity and suboptimal, inconsistent behaviors. The pla\-nning-as-descent framework leverages the increased state-space coverage present in noisy demonstrations to synthesize effective and generalizable goal-conditioned plans, rather than overfitting to specific behavioral patterns. PaD’s high and stable success rates suggest that its energy-based planning mechanism can extract useful structural information from widely varying suboptimal trajectories, enabling flexible recombination of behavioral fragments to achieve new goals. This robustness to demonstration noise highlights a key advantage of PaD over direct imitation methods, which often struggle when faced with inconsistent or low-quality supervision. PaD’s ability to generalize from such data further underscores the strength of its iterative refinement and verification-based synthesis approach.

\subsection{RQ3: Can Suboptimal Demonstrations Yield More Efficient Plans?}
Beyond success rates, we evaluate the efficiency of PaD's planning by measuring the average number of steps required to solve each task, comparing models trained on expert demonstrations (PaD-play) versus those trained on highly suboptimal behavior (PaD-noisy). For each of the five OGBench single-cube manipulation tasks, we record the episode lengths (i.e., the number of environment steps to reach the goal) over successful rollouts and report results in Table \ref{tab:avgsteps}.

\begin{table}[!htb]
\caption{\textbf{Average number of environment steps required to solve each task (lower is better).} PaD trained on highly suboptimal, diverse demonstrations (PaD-noisy) consistently achieves more efficient plans than PaD trained on narrow expert demonstrations (PaD-play), despite the lower quality of the training data. Results are reported as mean $\pm$ standard deviation over eight seeds.}
\label{tab:avgsteps}
\centering
\begin{tabular}{@{}llllllll@{}}
\toprule
Model-Data  & Task1 & Task2 & Task3 & Task4 & Task5 & Overall \\ \midrule
PaD-play    & 83$\pm$\small{10} & 73$\pm$\small{7} & 66$\pm$\small{5} & 83$\pm$\small{11} & 84$\pm$\small{7} & 78$\pm$\small{7}  \\
PaD-noisy   & \textbf{69$\pm$\small{8}} & \textbf{59$\pm$\small{4}} & \textbf{53$\pm$\small{7}} & \textbf{67$\pm$\small{7}} & \textbf{69$\pm$\small{8}} & \textbf{63$\pm$\small{6}} \\
\bottomrule
\end{tabular}
\end{table}

\paragraph{Results.} Counterintuitively, we find that PaD-noisy consistently solves tasks in fewer steps than the variant trained on expert demonstrations (PaD-play). Across all five tasks, PaD-noisy achieves goal completion with an overall average of 63$\pm$6 steps per episode, compared to 78$\pm$7 for PaD-play. On each individual task, PaD-noisy achieves lower mean episode lengths, with per-task averages ranging from 53$\pm$7 to 69$\pm$8, while PaD-play episode lengths range from 66$\pm$5 to 84$\pm$7. 

Importantly, these step counts are computed only over successful rollouts. The corresponding task success rates are reported separately: PaD-play success rates are given in Table~\ref{tab:play}, and PaD-noisy success rates are given in Table~\ref{tab:noisy}. Thus, the efficiency comparison in Table~\ref{tab:avgsteps} is conditioned on successful task completion and should be interpreted jointly with the success-rate results reported earlier.

\begin{figure}[!htb]
    \centering
    \includegraphics[width=0.6\textwidth]{"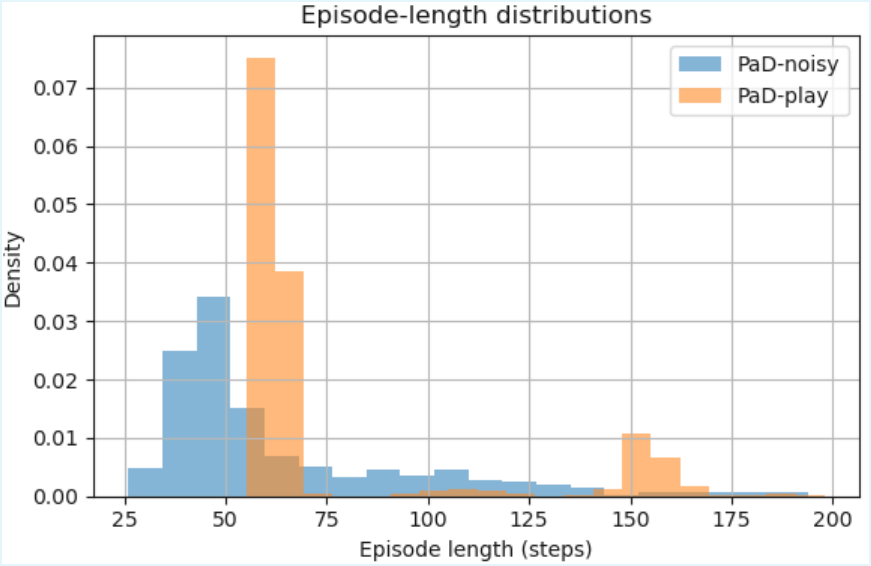"}
    \caption{\textbf{Episode-length distributions for cube-single tasks.} Histogram of episode lengths over successful rollouts for PaD-play and PaD-noisy, aggregated across all five cube-single manipulation tasks. PaD-noisy exhibits a clear left-shifted distribution, indicating consistently shorter plans compared to PaD-play. This confirms that the efficiency gains observed in mean episode length are systematic and not driven by outliers.}
    \label{fig:episode_lengths_plot}
\end{figure}

\paragraph{Episode-length distributions.}
To complement the mean $\pm$ std statistics in Table~\ref{tab:avgsteps}, Figure~\ref{fig:episode_lengths_plot} visualizes the full distribution of episode lengths over successful rollouts for both PaD-play and PaD-noisy, aggregated across all tasks. The distribution for PaD-noisy is clearly left-shifted relative to PaD-play, indicating that the reduction in average step count is not driven by a small number of unusually short episodes. Instead, PaD-noisy consistently produces shorter plans across the majority of successful rollouts, despite being trained on highly suboptimal and inconsistent demonstrations. In contrast, PaD-play exhibits a tighter but right-shifted distribution, closer to the demonstrator behavior, with a substantial fraction of episodes requiring more steps and a heavier tail corresponding to longer successful trajectories. This distributional view confirms that the efficiency gains observed for PaD-noisy reflect a systematic improvement in planning efficiency rather than an artifact of outliers or selective averaging.

\paragraph{Interpretation.} This counterintuitive result suggests that PaD is able to generalize more efficiently when trained on data with higher state-space coverage, even if the demonstrations themselves are highly suboptimal and inconsistent. The expert dataset appears to upper-bound the planner: since it only observes nearly optimal but narrowly distributed behaviors, PaD-play is restricted to reproducing these trajectories and struggles to discover more efficient solutions that are not present in the data. In contrast, the diverse transitions present in the noisy dataset allow PaD-noisy to observe a wider range of behaviors and transitions, enabling the planner to synthesize more efficient trajectories at test time--even outperforming the expert demonstrations themselves in terms of step efficiency. This finding highlights the strength of the planning-by-descent framework, which can leverage suboptimal and diverse demonstrations to achieve superior planning efficiency that goes beyond direct imitation, and underscores the importance of state-space diversity for enabling generalizable and efficient goal-reaching behavior.

\subsection{RQ4: What is the Impact of the Replanning Interval $N$ on Planning Robustness and Efficiency?}
\label{subsec:ablation-N}
\begin{figure}[!htb]
    \centering
    \includegraphics[width=\textwidth]{"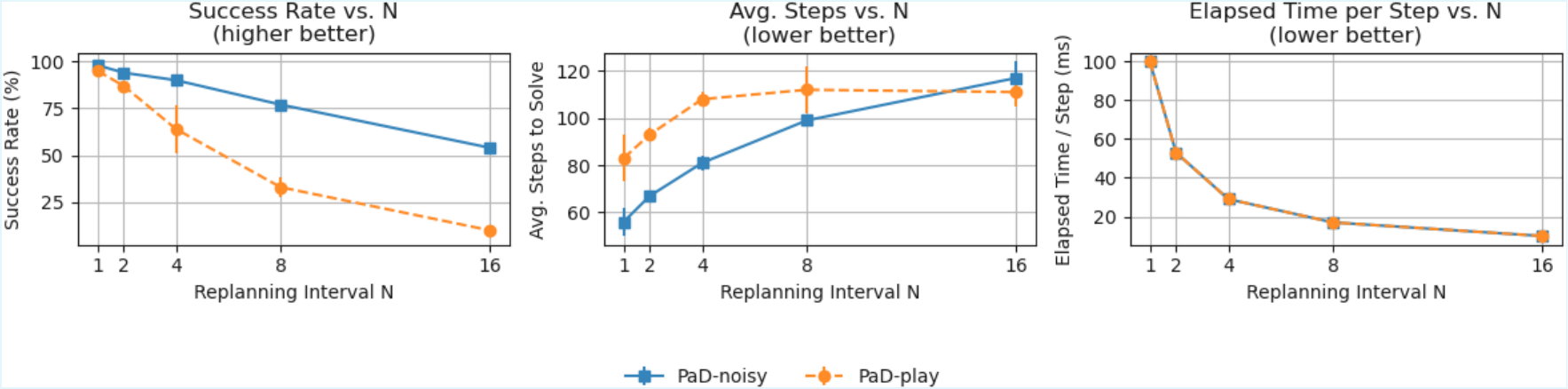"}
    \caption{\textbf{Effect of replanning interval $N$ on PaD performance.} Success rate, average steps to solve, and normalized computation time per step are shown for both PaD-noisy and PaD-play. PaD-noisy maintains higher success and efficiency at larger $N$, while PaD-play degrades rapidly, highlighting the greater robustness of planning from diverse demonstrations. Results are reported as mean $\pm$ standard deviation over eight seeds.}
    \label{fig:rq4}
\end{figure}

We analyze how varying the replanning interval $N$ affects both the robustness and efficiency of PaD when trained on narrow expert demonstrations (PaD-play) versus highly suboptimal, diverse demonstrations (PaD-noisy). Figure \ref{fig:rq4} summarizes the success rate, average steps to solve, and normalized computation time per environment step as $N$ increases from 1 (frequent replanning) to 16 (infrequent replanning).

Both variants exhibit a clear trade-off: increasing $N$ leads to substantial reductions in computational cost, with per-step time dropping from $100$ ms at $N=1$ to $10$ ms at $N=16$. However, this efficiency gain comes at the expense of task performance. For PaD-noisy, the success rate remains high for moderate $N$ values, decreasing from $98\pm2$ at $N=1$ to $90\pm 2$ at $N=4$, and only dropping sharply beyond $N=8$. The average steps to solve increase gradually, suggesting the model maintains efficient planning across a wider range of $N$.

In contrast, PaD-play is markedly less robust to infrequent replanning. Its success rate falls rapidly, dropping to $64\%$ at $N=4$ and reaching just $10\%$ at $N=16$. Similarly, average episode length increases from $83 \pm 10$ steps at $N=1$ to $111 \pm 6$ at $N=16$, with a significant performance collapse at moderate $N$. These trends indicate that PaD-play is highly sensitive to the replanning interval and requires frequent replanning to maintain reliable behavior.

Comparing the two, PaD-noisy consistently outperforms PaD-play at all values of $N$, exhibiting both higher success rates and greater resilience to longer open-loop horizons. The results suggest that exposure to diverse, suboptimal behavior during training enables PaD-noisy to recover from off-distribution states more effectively, making it less reliant on frequent replanning. In contrast, the narrow state coverage of PaD-play limits its ability to adapt when replanning is infrequent, leading to rapid degradation in both robustness and efficiency.

Taken together, these findings highlight the importance of training data diversity for planning robustness and suggest that moderate values of $N$ (such as $N=2$ or $N=4$) offer an effective balance between computational efficiency and reliable goal-reaching performance—particularly when models are trained on broad, suboptimal demonstrations.

\subsection{RQ5: What is the Impact of the Lightweight Manifold Projector on PaD's Performance?}
\label{sec:projector_ablation}

\begin{table}[!htb]
  \centering
  \caption{\textbf{Effect of the projector on planning performance.} Removing the projector $p_\theta$ causes a substantial drop in success rate on \texttt{single-cube-noisy-v0}, highlighting its importance for stable and effective trajectory refinement. Results are reported as mean $\pm$ standard deviation over four seeds.}
  \label{tab:projector_ablation}
  \small
  \begin{tabular}{l c}
    \toprule
    Method & Success Rate (\%) \\
    \midrule
    PaD w/o Projector $p_\theta$ & 50$\pm$2 \\
    PaD w/ Projector $p_\theta$  & \textbf{98$\pm$2} \\
    \bottomrule
  \end{tabular}
\end{table}

A key innovation in PaD is the introduction of a lightweight manifold projector network $p_\theta$, comprising just 130K parameters, which is applied after each gradient-based refinement step to ensure that latent trajectories remain close to the encoder-induced data manifold. To evaluate the necessity and effectiveness of this component, we train a version of PaD with the projector removed. 

Figure \ref{fig:projector_ablation} shows that removing the projector leads to degraded training dynamics, while Table \ref{tab:projector_ablation} confirms that it translates into a dramatic drop in final task performance: without the projector, success rates fall from above $95\%$ to just $50\%$. This large performance gap highlights the projector's crucial role in mediating between optimization and representation. By preventing energy descent from pushing latent states off-manifold, the projector helps maintain valid and dynamically consistent plans throughout refinement. Notably, this stabilization is achieved with minimal computational overhead, as the projector adds only a small fraction of the total model parameters. These results indicate that the manifold projector is not merely a lightweight architectural addition, but an essential component for robust and effective planning in the PaD framework.

\subsection{RQ6: How Sensitive is PaD to the Top-$K$ Selection Parameter?}

We assess the impact of the Top-$K$ selection parameter—which determines how many of the lowest-energy candidate trajectories are considered during plan selection—on PaD’s performance and efficiency. Across both PaD-noisy and PaD-play, we find that varying $K$ across a broad range ($K = {1, 5, 25, 50, 100}$, with $B=768$ candidates) has a negligible effect on success rate, average steps to solve, or computational cost. Performance remains statistically indistinguishable across all tested values of $K$. This insensitivity indicates that the energy landscape learned by PaD consistently yields robust, low-energy plans, rendering additional selection diversity largely unnecessary. Our findings underscore PaD’s robustness to the choice of $K$, thereby simplifying hyperparameter tuning and practical deployment.

\section{Discussion and Limitations}

This work introduces Planning as Descent (PaD), a framework that reframes offline goal-conditioned planning as gradient-based refinement in a learned energy landscape. By unifying trajectory evaluation and synthesis and enforcing training-inference alignment, PaD achieves strong and consistent performance across both narrow expert and highly suboptimal data regimes. The results suggest that learning to verify trajectories can provide a stable alternative to policy learning or sampling-based trajectory generation in offline settings where distribution shift and noisy demonstrations are prevalent.

Despite these strengths, several limitations remain. First, our experiments are conducted in a relatively modest computational regime, using lightweight models (6.5M parameters) and short planning horizons ($\sim10^2$ steps). While this demonstrates that PaD can be effective under limited resources, it prevents a systematic investigation of how the method scales with increased model capacity, longer horizons, and larger datasets. Understanding these scaling properties is an important direction for future work.

Second, we restrict our studies to state-based inputs and do not address learning directly from high-dimensional visual observations. Extending PaD to pixel-based or multimodal settings would require integrating visual representation learning in jointly learned latent spaces. Applying energy-based trajectory refinement in such settings remains an open and promising challenge that would significantly broaden the applicability of the framework.

Third, our empirical evaluation is limited to simulated manipulation tasks from the OGBench single-cube suite. While these tasks provide a controlled and challenging benchmark for offline goal-conditioned planning, they do not capture the full complexity of real-world robotic systems, including sensing noise, unmodeled dynamics, and actuation delays. Evaluating PaD on physical robotic platforms is a necessary step to assess the practical viability of energy-based planning-by-descent in real-world settings.

Finally, our experiments do not explore other important control domains such as conti\-nuous-control locomotion or navigation. Broadening empirical coverage to these domains would help clarify the generality and limitations of the planning-as-descent paradigm.

Overall, this work represents a step toward verification-driven planning frameworks that operate effectively from offline, reward-free data. We view PaD not as a final solution, but as a foundation for exploring how learned energy landscapes can support scalable, general-purpose planning and reasoning in increasingly complex environments.

\section{Conclusions}
We presented Planning as Descent (PaD), a framework for offline goal-conditioned planning that casts trajectory synthesis as gradient-based refinement in a learned energy landscape. By learning to verify entire future trajectories rather than directly generating them, and by enforcing alignment between training and inference through shared refinement dynamics, PaD provides a robust alternative to policy- and sampling-based approaches in reward-free, offline settings.

Empirically, PaD achieves strong performance on challenging OGBench single-cube manipulation tasks, including state-of-the-art results when trained on narrow expert demonstrations. Moreover, we observe that training on diverse but highly suboptimal data can further improve both success rates and planning efficiency. We interpret this effect as evidence that broad state-space coverage is particularly beneficial for verification-driven planning, enabling the energy landscape to support flexible trajectory refinement beyond the behaviors explicitly demonstrated in the data.

Taken together, these results suggest that learning to evaluate and refine trajectories may offer a scalable and principled foundation for offline goal-conditioned planning. We hope this work motivates further investigation into energy-based formulations as a unifying framework for planning, representation learning, and control.

% Acknowledgements and Disclosure of Funding should go at the end, before appendices and references

\acks{\paragraph{Funding.} This research was funded by the Valencian Institute for Business Competitiveness (IVACE) of Generalitat Valenciana}

\appendix

\section{Evaluation Protocol Details}
\label{app:evaluation}

OGBench defines a standardized evaluation protocol designed to enable robust and comparable measurements across offline goal-conditioned reinforcement learning algorithms \citep{park2024ogbench}. Under the official benchmark, agents are trained for up to 1M updates and evaluated every 100K steps using 50 rollouts for each of the five predefined test-time goals. Reported performance corresponds to the average success rate over the final three evaluation checkpoints (800K, 900K, and 1M updates for state-based tasks), resulting in a total of 750 evaluation episodes per agent (3 checkpoints $\times$ 5 goals $\times$ 50 rollouts).

All baseline results reported in this paper (GCBC, GCIVL, GCIQL, HIQL, CRL, QRL) are taken directly from the official OGBench benchmark and correspond to models trained and evaluated under this full protocol. Due to computational constraints, we do not retrain baseline methods under our reduced-budget setting.

For PaD, we adopt the same core structural components of the OGBench protocol—including multi-goal evaluation, randomized initial-state and goal perturbations, and multi-seed averaging—while using a substantially reduced training and evaluation budget. Specifically, PaD is trained for 200K gradient updates and evaluated once at the final checkpoint. For each of the five official test-time goals, we perform 20 evaluation episodes, resulting in 100 total rollouts per trained model. This reduction in rollout count follows the flexibility explicitly permitted by OGBench, which allows the use of fewer evaluation episodes per goal (e.g., 20 instead of 50) to reduce computational cost.

To account for training stochasticity, we train and evaluate eight independent PaD models with distinct random seeds, matching the multi-seed aggregation used for state-based OGBench baselines. Final performance for PaD is reported as the mean and standard deviation across these eight runs.

Overall, this evaluation protocol preserves the key comparability properties of OGBench--multi-goal evaluation, randomized rollouts, and multi-seed reporting--while making the comparison conservative with respect to PaD, as it is trained and evaluated under a substantially smaller computational budget than the reported baselines.

\section{Additional Implementation Details}

We provide additional architectural and optimization details to facilitate reproducibility.

\paragraph{Latent state representation.}
The state encoder maps each input state to a latent vector of dimension $d = 128$. Planning is performed over a fixed future horizon of $H = 80$ latent states. The maximum past context length is set to $P_{\max} = 16$.

\paragraph{Energy-based model architecture.}
The conditional energy model consists of a convolutional encoder followed by a decoder-only Transformer. The convolutional encoder uses 2 layers with kernel size 3 and channel dimensions ($d$, 256, 384) reducing the temporal resolution by a factor of 4. The Transformer comprises 3 blocks with 4 attention heads per block and a model dimension of 384.

\paragraph{Optimization and refinement.}
All networks are trained using the AdamW optimizer with a cosine decay learning rate from $3e^{-4}$ to $3e^{-5}$ and default momentum parameters. Gradient-based trajectory refinement uses a learnable step size $\eta$ initialized at $2.5$ and $T = 2$ refinement steps. Stop-gradient operations are applied between refinement steps as described in Section \ref{sec:method}.

\paragraph{Training details.}
Models are trained with batch size $512$ for 200K gradient updates. Unless otherwise stated, all hyperparameters are fixed across experiments and datasets.

\paragraph{Source code and reproducibility.} An implementation of the proposed method will be made publicly available at: \url{https://github.com/inescopresearch/pad}

\vskip 0.2in
\bibliography{sample}

@article{park2024ogbench,
  title={Ogbench: Benchmarking offline goal-conditioned rl},
  author={Park, Seohong and Frans, Kevin and Eysenbach, Benjamin and Levine, Sergey},
  journal={arXiv preprint arXiv:2410.20092},
  year={2024}
}

@inproceedings{lynch2020learning,
  title={Learning latent plans from play},
  author={Lynch, Corey and Khansari, Mohi and Xiao, Ted and Kumar, Vikash and Tompson, Jonathan and Levine, Sergey and Sermanet, Pierre},
  booktitle={Conference on robot learning},
  pages={1113--1132},
  year={2020},
  organization={Pmlr}
}

@article{park2023hiql,
  title={Hiql: Offline goal-conditioned rl with latent states as actions},
  author={Park, Seohong and Ghosh, Dibya and Eysenbach, Benjamin and Levine, Sergey},
  journal={Advances in Neural Information Processing Systems},
  volume={36},
  pages={34866--34891},
  year={2023}
}

@article{kostrikov2021offline,
  title={Offline reinforcement learning with implicit q-learning},
  author={Kostrikov, Ilya and Nair, Ashvin and Levine, Sergey},
  journal={arXiv preprint arXiv:2110.06169},
  year={2021}
}

@book{rawlings2020model,
  title={Model predictive control: theory, computation, and design},
  author={Rawlings, James Blake and Mayne, David Q and Diehl, Moritz and others},
  volume={2},
  year={2020},
  publisher={Nob Hill Publishing Madison, WI}
}

@inproceedings{wang2023optimal,
  title={Optimal goal-reaching reinforcement learning via quasimetric learning},
  author={Wang, Tongzhou and Torralba, Antonio and Isola, Phillip and Zhang, Amy},
  booktitle={International Conference on Machine Learning},
  pages={36411--36430},
  year={2023},
  organization={PMLR}
}

@article{schmidhuber2019reinforcement,
  title={Reinforcement Learning Upside Down: Don't Predict Rewards--Just Map Them to Actions},
  author={Schmidhuber, Juergen},
  journal={arXiv preprint arXiv:1912.02875},
  year={2019}
}

@article{eysenbach2022contrastive,
  title={Contrastive learning as goal-conditioned reinforcement learning},
  author={Eysenbach, Benjamin and Zhang, Tianjun and Levine, Sergey and Salakhutdinov, Russ R},
  journal={Advances in Neural Information Processing Systems},
  volume={35},
  pages={35603--35620},
  year={2022}
}

@article{andrychowicz2017hindsight,
  title={Hindsight experience replay},
  author={Andrychowicz, Marcin and Wolski, Filip and Ray, Alex and Schneider, Jonas and Fong, Rachel and Welinder, Peter and McGrew, Bob and Tobin, Josh and Pieter Abbeel, OpenAI and Zaremba, Wojciech},
  journal={Advances in neural information processing systems},
  volume={30},
  year={2017}
}

@article{fu2020d4rl,
  title={D4rl: Datasets for deep data-driven reinforcement learning},
  author={Fu, Justin and Kumar, Aviral and Nachum, Ofir and Tucker, George and Levine, Sergey},
  journal={arXiv preprint arXiv:2004.07219},
  year={2020}
}

@article{gladstone2025energy,
  title={Energy-Based Transformers are Scalable Learners and Thinkers},
  author={Gladstone, Alexi and Nanduru, Ganesh and Islam, Md Mofijul and Han, Peixuan and Ha, Hyeonjeong and Chadha, Aman and Du, Yilun and Ji, Heng and Li, Jundong and Iqbal, Tariq},
  journal={arXiv preprint arXiv:2507.02092},
  year={2025}
}

@article{park2025flow,
  title={Flow q-learning},
  author={Park, Seohong and Li, Qiyang and Levine, Sergey},
  journal={arXiv preprint arXiv:2502.02538},
  year={2025}
}

@article{lecun2006tutorial,
  title={A tutorial on energy-based learning},
  author={LeCun, Yann and Chopra, Sumit and Hadsell, Raia and Ranzato, M and Huang, Fujie and others},
  journal={Predicting structured data},
  volume={1},
  number={0},
  year={2006}
}

@article{lecun2022path,
  title={A path towards autonomous machine intelligence version 0.9. 2, 2022-06-27},
  author={LeCun, Yann},
  journal={Open Review},
  volume={62},
  number={1},
  pages={1--62},
  year={2022}
}

@article{levine2020offline,
  title={Offline reinforcement learning: Tutorial, review, and perspectives on open problems},
  author={Levine, Sergey and Kumar, Aviral and Tucker, George and Fu, Justin},
  journal={arXiv preprint arXiv:2005.01643},
  year={2020}
}

@article{west2023generative,
  title={The Generative AI paradox:" What it can create, it may not understand"},
  author={West, Peter and Lu, Ximing and Dziri, Nouha and Brahman, Faeze and Li, Linjie and Hwang, Jena D and Jiang, Liwei and Fisher, Jillian and Ravichander, Abhilasha and Chandu, Khyathi and others},
  journal={arXiv preprint arXiv:2311.00059},
  year={2023}
}

@article{janner2022planning,
  title={Planning with diffusion for flexible behavior synthesis},
  author={Janner, Michael and Du, Yilun and Tenenbaum, Joshua B and Levine, Sergey},
  journal={arXiv preprint arXiv:2205.09991},
  year={2022}
}

@article{chi2023diffusion,
  title={Diffusion policy: Visuomotor policy learning via action diffusion},
  author={Chi, Cheng and Xu, Zhenjia and Feng, Siyuan and Cousineau, Eric and Du, Yilun and Burchfiel, Benjamin and Tedrake, Russ and Song, Shuran},
  journal={The International Journal of Robotics Research},
  pages={02783649241273668},
  year={2023},
  publisher={SAGE Publications Sage UK: London, England}
}

@inproceedings{talvitie2017self,
  title={Self-correcting models for model-based reinforcement learning},
  author={Talvitie, Erik},
  booktitle={Proceedings of the AAAI conference on artificial intelligence},
  volume={31},
  number={1},
  year={2017}
}

@article{black2024pi_0,
  title={$\backslash pi\_0 $: A Vision-Language-Action Flow Model for General Robot Control},
  author={Black, Kevin and Brown, Noah and Driess, Danny and Esmail, Adnan and Equi, Michael and Finn, Chelsea and Fusai, Niccolo and Groom, Lachy and Hausman, Karol and Ichter, Brian and others},
  journal={arXiv preprint arXiv:2410.24164},
  year={2024}
}

@article{liu2024rdt,
  title={Rdt-1b: a diffusion foundation model for bimanual manipulation},
  author={Liu, Songming and Wu, Lingxuan and Li, Bangguo and Tan, Hengkai and Chen, Huayu and Wang, Zhengyi and Xu, Ke and Su, Hang and Zhu, Jun},
  journal={arXiv preprint arXiv:2410.07864},
  year={2024}
}

@article{balim2025model,
  title={Model-Based Diffusion Sampling for Predictive Control in Offline Decision Making},
  author={Balim, Haldun and Li, Na and Du, Yilun},
  journal={arXiv preprint arXiv:2512.08280},
  year={2025}
}

@article{bjorck2025gr00t,
  title={Gr00t n1: An open foundation model for generalist humanoid robots},
  author={Bjorck, Johan and Casta{\~n}eda, Fernando and Cherniadev, Nikita and Da, Xingye and Ding, Runyu and Fan, Linxi and Fang, Yu and Fox, Dieter and Hu, Fengyuan and Huang, Spencer and others},
  journal={arXiv preprint arXiv:2503.14734},
  year={2025}
}

@article{luo2024survey,
  title={A survey on model-based reinforcement learning},
  author={Luo, Fan-Ming and Xu, Tian and Lai, Hang and Chen, Xiong-Hui and Zhang, Weinan and Yu, Yang},
  journal={Science China Information Sciences},
  volume={67},
  number={2},
  pages={121101},
  year={2024},
  publisher={Springer}
}

@article{chen2021decision,
  title={Decision transformer: Reinforcement learning via sequence modeling},
  author={Chen, Lili and Lu, Kevin and Rajeswaran, Aravind and Lee, Kimin and Grover, Aditya and Laskin, Misha and Abbeel, Pieter and Srinivas, Aravind and Mordatch, Igor},
  journal={Advances in neural information processing systems},
  volume={34},
  pages={15084--15097},
  year={2021}
}

@inproceedings{wu2023masked,
  title={Masked trajectory models for prediction, representation, and control},
  author={Wu, Philipp and Majumdar, Arjun and Stone, Kevin and Lin, Yixin and Mordatch, Igor and Abbeel, Pieter and Rajeswaran, Aravind},
  booktitle={International Conference on Machine Learning},
  pages={37607--37623},
  year={2023},
  organization={PMLR}
}

@article{janner2021offline,
  title={Offline reinforcement learning as one big sequence modeling problem},
  author={Janner, Michael and Li, Qiyang and Levine, Sergey},
  journal={Advances in neural information processing systems},
  volume={34},
  pages={1273--1286},
  year={2021}
}

@article{sobal2025learning,
  title={Learning from reward-free offline data: A case for planning with latent dynamics models},
  author={Sobal, Vlad and Zhang, Wancong and Cho, Kyunghyun and Balestriero, Randall and Rudner, Tim GJ and LeCun, Yann},
  journal={arXiv preprint arXiv:2502.14819},
  year={2025}
}

@article{henaff2019model,
  title={Model-predictive policy learning with uncertainty regularization for driving in dense traffic},
  author={Henaff, Mikael and Canziani, Alfredo and LeCun, Yann},
  journal={arXiv preprint arXiv:1901.02705},
  year={2019}
}

@article{hansen2023td,
  title={Td-mpc2: Scalable, robust world models for continuous control},
  author={Hansen, Nicklas and Su, Hao and Wang, Xiaolong},
  journal={arXiv preprint arXiv:2310.16828},
  year={2023}
}

@article{zhou2024dino,
  title={Dino-wm: World models on pre-trained visual features enable zero-shot planning},
  author={Zhou, Gaoyue and Pan, Hengkai and LeCun, Yann and Pinto, Lerrel},
  journal={arXiv preprint arXiv:2411.04983},
  year={2024}
}

@article{wang2023energy,
  title={Energy-inspired self-supervised pretraining for vision models},
  author={Wang, Ze and Wang, Jiang and Liu, Zicheng and Qiu, Qiang},
  journal={arXiv preprint arXiv:2302.01384},
  year={2023}
}

@inproceedings{florence2022implicit,
  title={Implicit behavioral cloning},
  author={Florence, Pete and Lynch, Corey and Zeng, Andy and Ramirez, Oscar A and Wahid, Ayzaan and Downs, Laura and Wong, Adrian and Lee, Johnny and Mordatch, Igor and Tompson, Jonathan},
  booktitle={Conference on robot learning},
  pages={158--168},
  year={2022},
  organization={PMLR}
}

@article{liu2020energy,
  title={Energy-based imitation learning},
  author={Liu, Minghuan and He, Tairan and Xu, Minkai and Zhang, Weinan},
  journal={arXiv preprint arXiv:2004.09395},
  year={2020}
}

@article{carroll2022uni,
  title={Uni [mask]: Unified inference in sequential decision problems},
  author={Carroll, Micah and Paradise, Orr and Lin, Jessy and Georgescu, Raluca and Sun, Mingfei and Bignell, David and Milani, Stephanie and Hofmann, Katja and Hausknecht, Matthew and Dragan, Anca and others},
  journal={Advances in neural information processing systems},
  volume={35},
  pages={35365--35378},
  year={2022}
}

@article{du2019implicit,
  title={Implicit generation and modeling with energy based models},
  author={Du, Yilun and Mordatch, Igor},
  journal={Advances in neural information processing systems},
  volume={32},
  year={2019}
}

@article{ghosh2019learning,
  title={Learning to reach goals via iterated supervised learning},
  author={Ghosh, Dibya and Gupta, Abhishek and Reddy, Ashwin and Fu, Justin and Devin, Coline and Eysenbach, Benjamin and Levine, Sergey},
  journal={arXiv preprint arXiv:1912.06088},
  year={2019}
}

@article{davies2025ebt,
  title={EBT-Policy: Energy Unlocks Emergent Physical Reasoning Capabilities},
  author={Davies, Travis and Huang, Yiqi and Gladstone, Alexi and Liu, Yunxin and Chen, Xiang and Ji, Heng and Liu, Huxian and Hu, Luhui},
  journal={arXiv preprint arXiv:2510.27545},
  year={2025}
}

\end{document}